%% file: acl_main.tex
\pdfoutput=1
\documentclass[11pt]{article}

\usepackage[preprint]{acl}
\usepackage{times}
\usepackage{latexsym}

\usepackage{tikz}
\usepackage{graphicx}

\usetikzlibrary{positioning}
\usepackage{fontawesome5} % For user and bot icons

\usepackage{adjustbox}

\usepackage[T1]{fontenc}
\usepackage[utf8]{inputenc}

\usepackage{microtype}

\usepackage{inconsolata}

\usepackage{graphicx}

\title{ALN-P3: Unified Language Alignment for Perception, Prediction, and Planning in Autonomous Driving}

\author{
\normalfont{Yunsheng Ma$^{1,2}$ \ \  
Burhaneddin Yaman$^{1}$ \ 
Xin Ye$^{1}$\thanks{Corresponding author.}  \ 
Mahmut Yurt$^{1}$}\\  
Jingru Luo$^{1}$\
Abhirup Mallik$^{1}$ \ 
Ziran Wang$^{2}$ \ 
Liu Ren$^{1}$\\
$^{1}$Bosch Research North America \& Bosch Center for Artificial Intelligence (BCAI) \ \ \ \\ 
$^{2}$Purdue University\\
{\tt\small \{yunsheng,ziran\}@purdue.edu} \ \ \ \\  {\tt\small \{xin.ye3, jingru.luo, abhirup.mallik,liu.ren\}@us.bosch.com}
}
\input{preamble}
\begin{document}
\maketitle
\input{sec/0_abstract}

\input{sec/1_intro}
\input{sec/2_related_work}

\input{sec/3_methodology}

\input{sec/4_experiments}
\input{sec/5_conclusion}

\bibliography{custom,ma,ma2}

% \appendix

% \section{Example Appendix}
% \label{sec:appendix}

% This is an appendix.

\end{document}

%% file: preamble.tex
\usepackage{multirow}
\usepackage{arydshln}
\usepackage{amsmath,amssymb,amsfonts}
\usepackage{gensymb} % degree
\usepackage{booktabs}

\newcommand{\NAME}{ALN-P3 }

\usepackage[capitalize]{cleveref}
\crefname{section}{Sec.}{Secs.}
\Crefname{section}{Section}{Sections}
\Crefname{table}{Table}{Tables}
\crefname{table}{Tab.}{Tabs.}

%% file: sec/0_abstract.tex
\begin{abstract}
Recent advances have explored integrating large language models (LLMs) into end-to-end autonomous driving systems to enhance generalization and interpretability. However, most existing approaches are limited to either driving performance or vision-language reasoning, making it difficult to achieve both simultaneously. In this paper, we propose \textbf{ALN-P3}, a unified co-distillation framework that introduces cross-modal alignment between "fast" vision-based autonomous driving systems and "slow" language-driven reasoning modules. ALN-P3 incorporates three novel alignment mechanisms: Perception Alignment (P1A), Prediction Alignment (P2A), and Planning Alignment (P3A), which explicitly align visual tokens with corresponding linguistic outputs across the full perception, prediction, and planning stack. All alignment modules are applied only during training and incur no additional costs during inference. Extensive experiments on four challenging benchmarks—nuScenes, Nu-X, TOD3Cap, and nuScenes QA—demonstrate that ALN-P3 significantly improves both driving decisions and language reasoning, achieving state-of-the-art results.
\end{abstract}

%% file: sec/1_intro.tex
\section{Introduction}
\label{sec:intro}
End-to-end autonomous driving systems, which integrate perception, prediction, and planning (collectively referred to as the P3 stack), have made significant progress in recent years~\cite{hu_st-p3_2022,hu_planning-oriented_2023, jiang_vad_2023, weng_para-drive_2024,jia_drivetransformer_2025}.
These systems are often characterized as ``fast" due to their low-latency performance, making them suitable for real-time deployment in real-world environments. However, they face notable limitations in reasoning, generalizability, and interpretability, as they typically operate solely on multi-view camera inputs and lack understanding based on commonsense knowledge.

To address these shortcomings, recent research has explored the use of foundation models, such as large language models (LLMs) and vision-language models (VLMs), within autonomous driving pipelines~\cite{hwang_emma_2024,sima_drivelm_2024,ma_position_2025,renz_simlingo_2025,wang_omnidrive_2025}. These models incorporate both visual and linguistic information to perform P3 tasks and offer improved capabilities for reasoning and explanation~\cite{hwang_emma_2024}. Nonetheless, these models are often referred to as "slow" systems due to their long inference time, which remains a major obstacle to real-world deployment.

Another important direction in autonomous driving research involves the development of "slow-and-fast" systems~\cite{tian_drivevlm_2024}. In this paradigm, the "fast" system handles core P3 tasks such as planning, while the "slow" system operates in parallel to provide interpretability and reasoning for the actions taken. Although this architecture offers a promising solution to long-standing challenges in explainability, the interaction between the two systems has not been fully exploited to benefit both streams.

In existing works, the pretrained features from the "fast" system are simply propagated to the "slow" system to enhance its language output~\cite{ding_hint-ad_2024}. However, there has been little focus on explicitly aligning the perception, prediction, and planning components of the "fast" system with the reasoning of the "slow" system. As a result, current "slow-and-fast" systems often suffer from inconsistency between the two branches, leading to suboptimal performance in real-world autonomous driving systems.

To address these challenges, we introduce \NAME, a novel framework that integrates vision and language alignment across the entire autonomous driving stack. ALN-P3 is a training-only co-distillation strategy that aligns the "slow" and "fast" systems to enhance the performance of both, without introducing any additional computational cost to the "fast" system at inference time. This design makes ALN-P3 practical for real-world deployment, where low-latency decision making is critical for practicality.

Specifically, ALN-P3 introduces three alignment modules: Perception Alignment (P1A), Prediction Alignment (P2A), and Planning Alignment (P3A). They connect BEV-based visual representations from the "fast" system with their corresponding language-based representations from the "slow" system. Specifically, P1A aligns instance-level BEV features with object-level captions, P2A aligns predicted motion trajectories with agent-centric language outputs, and finally P3A aligns ego vehicle plans with planning-related natural language responses.

All alignment modules are applied only during training and do not add any computational cost during inference. This makes ALN P3 suitable for real-time deployment in autonomous driving systems. We conduct an extensive evaluation study on four challenging benchmarks: nuScenes, Nu-X, TOD3Cap, and nuScenes QA. \NAME consistently improves both planning performance in the "fast" system and reasoning accuracy in the "slow" system. For instance, it reduces the average collision rate by 27\% compared to prior planners and achieves a 28\% improvement in CIDEr score on the Nu-X driving explanation task.

Our main contributions are summarized as follows:
\begin{itemize}
\item We propose \textbf{ALN-P3}, a novel co-distillation framework that introduces alignment between the "slow" and "fast" systems across the entire perception, prediction, and planning (P3) stack in autonomous driving.
\item \textbf{ALN-P3} incorporates three alignment modules—Perception Alignment (P1A), Prediction Alignment (P2A), and Planning Alignment (P3A)—to explicitly align visual and linguistic representations between the two systems for each core driving task.
\item \textbf{ALN-P3} is a training-only approach that incurs no additional computational cost during inference, making it efficient and suitable for real-time deployment in autonomous driving systems.
\item Extensive experiments on four challenging benchmarks demonstrate that \textbf{ALN-P3} consistently outperforms existing methods on both planning and reasoning tasks, validating the effectiveness of the proposed cross-system co-distillation strategy.
\end{itemize}

%% file: sec/2_related_work.tex
\section{Related Work}
\label{sec:related_work}

\subsection{End-to-End Autonomous Driving}
Autonomous driving systems consist of perception, prediction and planning tasks (P3). While these tasks were traditionally trained and optimized separately, end-to-end systems that train and optimize all three tasks jointly have gained more interest as they show significant performance improvements \cite{chen_end--end_2024,hu_planning-oriented_2023,jiang_vad_2023,li_is_2024,weng_para-drive_2024}. These vision-only approaches are generally referred as "fast" systems due to their low-latency inference capabilities~\cite{tian_drivevlm_2024}.

Among these works, UniAD has played a pivotal role in the rapid emergence and adoption of end-to-end autonomous driving systems. Specifically, UniAD employs rasterized representations and performs planning-oriented optimization for end-to-end autonomous driving~\cite{hu_planning-oriented_2023}. Subsequently, VAD proposed a vectorized representation to reduce computational costs~\cite{jiang_vad_2023}. In another line of work, PARA-Drive and DriveTransformer introduced parallel architectures to further enhance the efficiency of end-to-end autonomous driving systems~\cite{weng_para-drive_2024,jia_drivetransformer_2025}.

\subsection{VLMs for Autonomous Driving}
Vision-language models (VLMs) have emerged as an alternative approach for end-to-end autonomous driving. These models leverage both visual and linguistic information and are known as "slow" systems due to their longer inference times~\cite{tian_drivevlm_2024}.. Notably, EMMA~\cite{hwang_emma_2024}, developed by Waymo, fine-tunes  Gemini~\cite{google_gemini_2024} multimodal large language model for autonomous driving tasks. EMMA processes raw camera inputs and textual data to generate various driving outputs, including planner trajectories, perception objects, and road graph elements. By representing non-sensor inputs and outputs as natural language text, EMMA maximizes the utility of world knowledge from pre-trained large language models, allowing it to jointly process various driving tasks in a unified language space.

Other notable VLM-based autonomous driving models include DriveVLM~\cite{tian_drivevlm_2024}, DiMA~\cite{hegde_distilling_2025}, SimLingo~\cite{renz_simlingo_2025}, OmniDrive~\cite{wang_omnidrive_2025}, and TOKEN~\cite{tian_tokenize_2024}. These models explore different strategies for integrating visual and linguistic information to enhance driving performance and reasoning capabilities.

\subsection{Alignment Mechanisms}
Alignment mechanisms have been proposed to bridge the gap between different modalities in multimodal models~\cite{chen_internvl_2024,radford_learning_2021,khattak_maple_2023,yu_representation_2025}. CLIP~\cite{radford_learning_2021} introduced a contrastive learning approach to align image and text representations in a shared embedding space. REPresentation Alignment (REPA) is a regularization technique designed to enhance the efficiency and quality of training in diffusion models by aligning the projections of noisy input hidden states in denoising networks with clean image representations obtained from external, pretrained visual encoders~\cite{yu_representation_2025}. REPA has demonstrated significant improvements in both training efficiency and generation quality when applied to popular diffusion and flow-based transformers.

These alignment techniques inform our proposed alignment modules (P1A, P2A, P3A) in ALN-P3, which aim to explicitly align visual tokens with corresponding linguistic outputs across the full perception, prediction, and planning stack.

%% file: sec/3_methodology.tex
\section{Methodology}
\newcommand{\qins}{\textbf{Q}'_\text{instance}}

\label{sec:methodology}
\begin{figure*}
    \centering
    \includegraphics[width=1.0\linewidth]{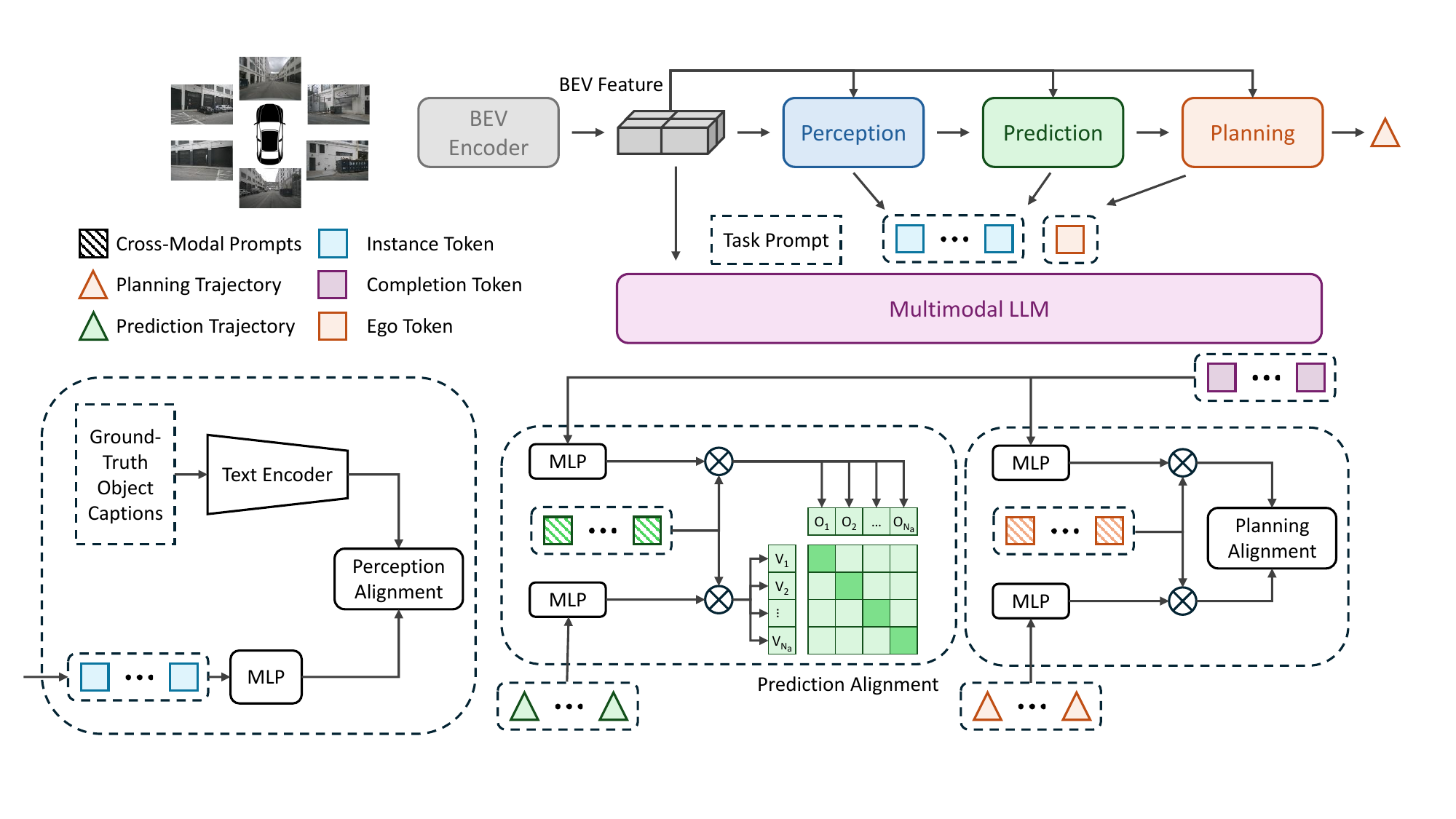}
    \caption{\textbf{Overview of the proposed \NAME framework}, which integrates vision and language alignment across the full autonomous driving stack. The architecture includes three alignment modules: Perception Alignment (P1A), Prediction Alignment (P2A), and Planning Alignment (P3A). These modules align BEV-based visual tokens, such as instance, motion, and ego features, with corresponding natural language outputs through cross-modal prompts and projection heads. All alignments are applied only during training and introduce no additional computation at inference time, enabling efficient and interpretable reasoning for perception, prediction, and planning tasks.}
    \label{fig:p3}

\end{figure*}
\textbf{Overview.} The overall framework of \NAME introduces multimodal alignment across the full autonomous driving stack—including perception, prediction, and planning—to improve both driving performance and language reasoning. An overview of the framework is illustrated in \cref{fig:p3}. The methodology section is organized as follows. In \cref{sec:prelim}, we introduce the P3 autonomous driving stack and the QA module. In \cref{sec:p1a}, \cref{sec:p2a}, and \cref{sec:p3a}, we present the proposed alignment modules: Perception Alignment (P1A), Prediction Alignment (P2A), and Planning Alignment (P3A).

\subsection{Preliminaries}
\label{sec:prelim}
\paragraph{P3 Module.}
The autonomous driving stack $\mathcal{D}$ follows a vision-based perception–prediction–planning (P3) pipeline~\cite{hu_st-p3_2022,hu_planning-oriented_2023}. Given multi-view image inputs, a BEV encoder extracts a bird’s-eye-view (BEV) feature map $\mathbf{B}$ by fusing multi-camera information into a unified spatial-temporal representation~\cite{li_bevformer_2022}.

In the perception stage, learnable track tokens $\mathbf{Q}_\text{track} \in \mathbb{R}^{N_a \times D_q}$ query the BEV feature map to detect and track agents, where $N_a$ denotes the number of agent queries and $D_q$ is the dimension of the query token. These tokens encode positional historical trajectory features of surrounding objects.

The prediction stage introduces motion tokens $\mathbf{Q}_\text{motion} \in \mathbb{R}^{N_a \times D_q}$ to model agent-agent and agent-map interactions, forecasting future trajectories $\mathbf{V}_a$ for each agent. In the planning stage, an ego-vehicle token $\mathbf{Q}_\text{ego}$ is used to infer the ego vehicle’s future trajectory $\mathbf{V}_\text{ego}$.

\paragraph{QA Module.}
The QA module $\mathcal{G}$ generates natural language responses to various driving tasks, conditioned on task-specific prompts~\cite{hwang_emma_2024}. Since LLMs cannot directly ingest visual features, a middleware encoder $\mathcal{E}$, such as a Q-Former~\cite{li_blip-2_2023}, is commonly employed to project visual representations into the language space. Specifically, we adopt the Holistic Token Mixer~\cite{ding_hint-ad_2024} to project the driving context into the MLLM space:
\begin{multline}
\textbf{C} = \left(\textbf{B}', \qins, \textbf{Q}'_\text{ego}, {\textbf{V}'}_\text{plan}\right) \\
= \mathcal{E}\left(\textbf{B}, \textbf{Q}_\text{track}, \textbf{Q}_\text{motion}, \textbf{Q}_\text{ego}, {\textbf{V}}_\text{plan}\right),
\end{multline}
where track and motion tokens are fused into instance-level representations that encapsulate both historical and predicted trajectories for each surrounding agent.

Given a task prompt $\mathbf{P}$ and the multimodal driving context $\mathbf{C}$, the MLLM generates a language output sequence $\mathbf{O}$ via an autoregressive decoder:
\begin{equation}
\Pr(\mathbf{O}|\mathbf{P}, \mathbf{C}) = \prod_{i=1}^{n} \Pr(o_i | o_{<i}, \mathbf{P}, \mathbf{C}),
\end{equation}
where $o_{<i}$ denotes the sequence of previously generated tokens. The QA module is trained using a standard language modeling loss.

\subsection{Perception Alignment}
\label{sec:p1a}

The goal of perception alignment is to bridge the gap between the scene representations used in the P3 Module and the language-based reasoning capabilities of MLLMs. Off-the-shelf MLLMs, however, are not trained to process driving-specific visual representations, making it difficult for them to comprehend or reason over such inputs. This misalignment is particularly pronounced when compared to visual tokens from general-domain MLLMs~\cite{chen_internvl_2024}.

To address this challenge, we propose a perception alignment module (P1A) that explicitly aligns the instance-level perception features—capturing visual and contextual information about each object—with their corresponding ground-truth linguistic descriptions. This alignment helps the MLLM better understand and reason over perception-level semantics.

Formally, given a set of ground-truth object captions $\mathbf{O}$, we compute their text embeddings using a pretrained CLIP text encoder $\mathcal{T}$~\cite{radford_learning_2021}. The instance-level visual features $\qins$ are projected into the language space via a trainable MLP head $\Phi^{P_1}(\cdot)$. We then enforce alignment using a mean squared error (MSE) loss:

\begin{equation}
\mathcal{L}_{\text{P1A}} = \left\|\Phi^{P_1}(\qins) - \mathcal{T}(\mathbf{O})\right\|_{2}^{2}.
\end{equation}

This alignment loss encourages the model to produce perception features that are semantically consistent with their linguistic counterparts, facilitating more effective cross-modal understanding.

\subsection{Prediction Alignment}
\label{sec:p2a} 
In current LLM-integrated autonomous driving systems, driving performance and vision-language reasoning are often optimized independently. This decoupling can result in suboptimal outcomes and inconsistencies between the vision-based trajectory predictions and the language outputs (e.g., captions or answers) generated by the MLLM. To address this issue, we propose a prediction alignment module (P2A) that explicitly bridges the gap between the predicted motion trajectories and the linguistic outputs.

The primary challenge lies in the heterogeneous nature of these modalities: trajectory predictions consist of continuous-valued spatial waypoints, whereas language outputs are represented as discrete token logits. Naively aligning these modalities can degrade the performance of both the prediction and generation tasks. To overcome this, we adopt a cross-modal prompting strategy.

We introduce a set of $N_2$ learnable prompt tokens $\mathcal{P}_2 \in \mathbb{R}^{N_2 \times D_2}$ that define a shared embedding space for aligning prediction and language features. Let $\mathbf{V}_a^k \in \mathbb{R}^{T_f \times 2}$ denote the predicted future trajectory of the $k$-th agent from the P3 module, where $T_f$ is the number of future time steps.

To map these modalities into the shared space, we apply an attention-based pooling $\mathcal{A}$, defined as:
\begin{multline}
\mathcal{A}(\mathcal{W}, \mathbf{H}) =\\ \sum_{i=1}^{N}
\left(\frac{ \exp \left( \Phi^{(\cdot)}(\mathbf{H})^\top \cdot \mathbf{w}_i \right) }{ \sum_{j=1}^{N} \exp \left( \Phi^{(\cdot)}(\mathbf{H})^\top \cdot \mathbf{w}_j \right) } \right) \mathbf{w}_i,
\end{multline}
where $\mathcal{W} = \left\{\mathbf{w}_1, \ldots, \mathbf{w}_N\right\} \in \mathbb{R}^{N \times D}$ is a set of prompt tokens, and $\Phi^{(\cdot)}$ is a trainable projection head (MLP) that transforms input features into the shared space.

Using this operator, we compute the aligned representations:
\begin{equation}
\mathbf{z}_{\text{pred}} = \mathcal{A}(\mathcal{P}_2, \mathbf{V}_a), \quad \mathbf{z}_{\text{llm}} = \mathcal{A}(\mathcal{P}_2, \mathbf{O}),
\end{equation}
where $\mathbf{O}$ denotes the output logits from the MLLM.

To align these embeddings, we employ a CLIP-style contrastive loss~\cite{radford_learning_2021}:
\begin{equation}
\mathcal{L}_{\text{P2A}}(\mathcal{D}, \mathcal{G}) = \mathcal{L}_{\text{CLIP}}(\mathbf{z}_{\text{pred}}, \mathbf{z}_{\text{llm}}).
\end{equation}

Here, both $\mathbf{z}_{\text{pred}}, \mathbf{z}_{\text{llm}} \in \mathbb{R}^{N_a \times D_2}$ are computed for each of the $N_a$ agents in a scene, enabling fine-grained agent-level alignment between predicted motion and corresponding language descriptions.

By minimizing $\mathcal{L}_{\text{P2A}}$, the P2A module encourages the trajectory and language branches to converge in the shared embedding space, promoting cross-modal consistency and reducing semantic drift.

% This module complements the perception alignment module (P1A) by addressing temporal consistency and future intent. Together, P1A and P2A offer holistic supervision for aligning visual perception, motion prediction, and language reasoning within end-to-end autonomous driving systems.

\subsection{Planning Alignment}
\label{sec:p3a}
While the Prediction Alignment module (P2A) promotes consistency between agent-level motion forecasting and linguistic reasoning, it does not directly supervise the decision-making process of the ego vehicle, which is the most critical aspect of autonomous driving. To address this gap, we introduce the Planning Alignment module (P3A), which aligns the ego vehicle’s planned trajectory with the language outputs generated in response to planning-related queries.

This alignment is grounded in the projected ego-vehicle's future trajectory $\mathbf{V}_\text{ego}$ produced by the P3 module. To bridge the modality gap between the planned trajectory and the corresponding language output, both are projected into a shared representation space using a set of learnable prompt tokens $\mathcal{P}_3 \in \mathbb{R}^{N_3 \times D_3}$.

Using the attention-based pooling operator $\mathcal{A}$, we obtain:
\begin{equation}
\mathbf{z}_{\text{plan}} = \mathcal{A}(\mathcal{P}_3, \mathbf{V}_\text{ego}), \quad \mathbf{z}_{\text{llm}} = \mathcal{A}(\mathcal{P}_3, \mathbf{O}),
\end{equation}
where $\mathbf{O}$ represents the MLLM’s output logits in response to planning-related prompts.

To align these two representations, we adopt a negative cosine similarity loss, an empirically effective objective for alignment~\cite{yu_representation_2025}:
\begin{equation}
\mathcal{L}_{\text{P3A}}(\mathcal{D}, \mathcal{G}) = -\frac{\mathbf{z}_{\text{plan}}^\top \cdot \mathbf{z}_{\text{llm}}}{\|\mathbf{z}_{\text{plan}}\|_2^2 \cdot \|\mathbf{z}_{\text{llm}}\|_2^2}.
\end{equation}

Since there is only one ego vehicle per scene, this loss is computed at the scene level. Minimizing $\mathcal{L}_{\text{P3A}}$ encourages the semantic embedding of the ego vehicle’s planned trajectory to be aligned with its language-based description, thereby enhancing consistency between the P3 module's decision output and QA module generated explanation or command.

Together with the perception (P1A) and prediction (P2A) alignment modules, P3A completes the supervision of the entire perception–prediction–planning (P3) stack. This comprehensive alignment fosters coherent, interpretable, and semantically grounded reasoning across all stages of the autonomous driving pipeline.

\textit{Importantly, all three alignment modules—P1A, P2A, and P3A—are only active during training and do not introduce any additional computational overhead at inference time, which is critical for real-world autonomous driving applications, where low-latency decision-making is essential}. Furthermore, during training, the specific alignment strategy to activate is determined dynamically based on the category of the prompt question in the training data. The prompt categories—perception, prediction, and planning—are defined following the taxonomy proposed in DriveLM~\cite{sima_drivelm_2024}.

%% file: sec/4_experiments.tex
\begin{table*}[!th]
    \centering
    \resizebox{\linewidth}{!}{
    \begin{tabular}{l|cccc|cccc|cccc|ccc}
        \hline
        \multirow{2}{*}{Method} & \multicolumn{4}{c|}{Collision (\%)} & \multicolumn{4}{c|}{Nu-X} & \multicolumn{4}{c|}{TOD3Cap} & \multicolumn{3}{c}{NuScenes-QA}\\
        & $1s$ & $2s$ & $3s$ & Avg. & C & B & M & R & C & B & M & R & H0 & H1 & All\\
        \hline
        ST-P3~\cite{hu_st-p3_2022} & 0.23 & 0.62 & 1.27 & 0.71 & - & - & - & - & - & - & - & - & - & - & - \\
        UniAD~\cite{hu_planning-oriented_2023} & \textbf{0.05} & \underline{0.17} & 0.71 & 0.31 & - & - & - & - & - & - & - & - & - & - & - \\
        VAD~\cite{jiang_vad_2023} & 0.07 & \underline{0.17} & 0.41 & \underline{0.22} & - & - & - & - & - & - & - & - & - & - & - \\
        DriveVLM~\cite{tian_drivevlm_2024} & 0.10 & 0.22 & \underline{0.45} & 0.27 & - & - & - & - & - & - & - & - & - & - & - \\
        TOD3Cap~\cite{jin_tod3cap_2024} & - &- & - & - & 14.5 & 2.45 & 10.5 & 23.0 & 120.3 & 51.5 & 45.1 & 70.1 & 53.0 & 45.1 & 49.0\\
        GPT-4o~\cite{openai_gpt-4_2023}$^\dagger$ & - & - & - & - & 19.0 & 3.95 & 10.3 & 24.9 & 160.8 & 50.4 & 31.6 & 43.5 & 42.0 & 34.7 & 37.1\\
        Gemini-1.5~\cite{google_gemini_2024}$^\dagger$ & - &- & - & - & 17.6 & 3.43 & 9.3 & 23.4 & 169.7 & 53.6 & 33.4 & 45.9 & 40.5 & 32.9 & 35.4\\
        Hint-AD~\cite{ding_hint-ad_2024} & 0.26 & 0.34 & 0.63 & 0.41 & 22.4 & 4.18 & 13.2 & 27.6 & 263.7 & 67.6 & \textbf{47.5} & 79.4 & 55.4 & 48.0 & 50.5\\
        \textbf{P3 (Ours)} & \textbf{0.05} & \textbf{0.09} & \textbf{0.35} & \textbf{0.16} & \textbf{28.6} & \textbf{5.59} & \textbf{14.7} & \textbf{35.2} & \textbf{341.9} & \textbf{70.8} & \underline{45.8} & \textbf{82.5} & \textbf{57.1} & \textbf{50.9} & \textbf{52.9}\\
        \hline 
    \end{tabular}
    }
    \caption{Quantitative comparison across four tasks: end-to-end planning (collision rate on nuScenes), driving explanation (Nu-X), 3D dense captioning (TOD3Cap), and visual question answering (nuScenes-QA). \NAME outperforms all baselines, achieving state-of-the-art results in both driving safety and multimodal reasoning. Metrics include collision rate (lower is better), and standard language metrics: CIDEr (C), BLEU-4 (B), METEOR (M), ROUGE-L (R), and QA accuracy (H0, H1, All).}
    \label{tab:main-result}
\end{table*}

\section{Experiments and Results}
\label{sec:experiments}
\subsection{Datasets and Metrics}
\label{sec:datasets}
We evaluate the proposed \NAME{} framework and baseline models on four diverse datasets to ensure a comprehensive and fair comparison across driving and language tasks.

\textbf{nuScenes}~\cite{caesar_nuscenes_2020} is a widely adopted benchmark for autonomous driving. It contains 1,000 20-second driving scenes, annotated at 2 Hz, with 1.4 million 3D bounding boxes captured from six cameras covering a full $360\degree$ field of view. We use the \emph{collision rate} at 1s, 2s, and 3s prediction horizons to evaluate end-to-end planning performance.

\textbf{Nu-X}~\cite{ding_hint-ad_2024} is a large-scale, human-annotated driving explanation dataset built upon nuScenes. It includes 34,000 keyframes with fine-grained narrations and reasoning, capturing both ``what" and ``why" aspects of driving behavior. We evaluate generated captions using standard captioning metrics: \emph{CIDEr} (C)~\cite{vedantam_cider_2015}, \emph{BLEU-4} (B)~\cite{papineni_bleu_2002}, \emph{METEOR} (M)~\cite{banerjee_meteor_2005}, and \emph{ROUGE-L} (R)~\cite{lin_rouge_2004} , which collectively assess fluency, relevance, and diversity.

\textbf{TOD3Cap}~\cite{jin_tod3cap_2024} is a 3D dense captioning dataset that provides object-level textual descriptions, covering attributes such as appearance, motion, and spatial relationships. Evaluation is conducted using the same captioning metrics as Nu-X, allowing us to measure grounding performance at the object level.

\textbf{nuScenes-QA}~\cite{qian_nuscenes-qa_2024} is a visual question answering dataset built on 34,000 nuScenes key frames. It includes five question types: existence, counting, query-object, query-status, and comparison. We report \emph{accuracy} across zero-hop, one-hop, and all question types, which reflect the model’s reasoning capabilities over visual contexts.

\subsection{Baselines}
\label{sec:baselines}
To benchmark the effectiveness of the proposed \NAME framework, we compare against a diverse set of baselines, including traditional autonomous driving pipelines, vision-language models, and state-of-the-art multimodal LLMs. Each baseline is evaluated across the four target tasks described in \cref{sec:datasets}.

\textbf{ST-P3}~\cite{hu_st-p3_2022}, \textbf{UniAD}~\cite{hu_planning-oriented_2023}, and \textbf{VAD}~\cite{jiang_vad_2023} are leading end-to-end vision-based perception–prediction–planning (P3) models. These models predict ego vehicle trajectories from multi-view inputs and serve as strong baselines for the planning task on nuScenes. However, they do not support language output and thus are not evaluated on reasoning, captioning, or VQA tasks.

\textbf{DriveVLM}~\cite{tian_drivevlm_2024} proposes a hybrid framework that integrates a vision-language model with a conventional AD stack. It provides language reasoning capability over driving context but lacks fine-grained alignment with intermediate outputs.

\textbf{TOD3Cap}~\cite{jin_tod3cap_2024} is a 3D dense captioning model that combines BEV-based detection with an LLM decoder to produce object-centric descriptions. However, the language output is not explicitly aligned with motion or planning representations, limiting its interpretability for downstream tasks.

\textbf{GPT-4o} \cite{openai_gpt-4_2023} and \textbf{Gemini-1.5} \cite{google_gemini_2024} are proprietary multimodal LLMs. We evaluate them in a zero-shot setting using multi-view camera images and manually constructed prompts. Although they offer strong general-purpose reasoning, they operate independently of any AD stack.

\textbf{Hint-AD}~\cite{ding_hint-ad_2024} incorporates intermediate outputs from an AD model into a LLM decoder using a holistic token mixer. It supports multiple reasoning tasks and offers stronger grounding than declarative baselines, but does not provide explicit alignment across all three P3 stages.

\subsection{Implementation Details}
For fair comparison, we follow the same experimental setup as Hint-AD~\cite{ding_hint-ad_2024}. ALN-P3 adopts the same base architecture, with the difference being the integration of our proposed alignment modules (P1A, P2A, P3A). We use VAD-base~\cite{jiang_vad_2023} as the pretrained P3 module. For the language generator, we employ LLaMA-AdapterV2~\cite{zhang_llama-adapter_2024}, built on top of the pretrained LLaMA-2-7B~\cite{touvron_llama_2023}.

All models are trained for 5 epochs on a mixed dataset comprising all four datasets. The base learning rate is set to $2 \times 10^{-4}$. During training, the LLaMA model is kept frozen. The adapter and P3 parameters are updated. All experiments are conducted on two NVIDIA H200 GPUs.

We apply all the training objectives across the P3 stack, together with the language modeling loss and all proposed alignment losses. All loss terms are equally weighted with a default weight of 1, and no additional tuning is applied to balance these terms due to resource constraints.

\subsection{Experiment Results}
\label{sec:results}
We report comprehensive results across four tasks: end-to-end planning (nuScenes), driving explanation and reasoning (Nu-X), 3D dense captioning (TOD3Cap), and visual question answering (nuScenes-QA). \cref{tab:main-result} presents quantitative comparisons with state-of-the-art baselines and multimodal LLMs.

\subsubsection{End-to-End Planning}
We evaluate planning safety using the collision rate at 1s, 2s, and 3s horizons on the nuScenes dataset. As shown in \cref{tab:main-result}, \NAME achieves the lowest collision rates across all time horizons (0.05\%, 0.09\%, and 0.35\%), outperforming all baselines, including strong vision-based planners like UniAD and VAD. Notably, our model achieves an average collision rate of 0.16\%, representing a 27\% improvement over the best prior model (VAD, 0.22\%). This demonstrates that introducing multimodal alignment can enhance low-level planning performance.

\subsubsection{Driving Explanation and Reasoning}
\NAME delivers state-of-the-art results on the Nu-X dataset, which evaluates language reasoning in autonomous driving. Compared to Hint-AD, our model improves CIDEr from 22.4 to 28.6, and BLEU-4 from 4.18 to 5.59. These gains indicate more informative, fluent, and contextually grounded driving explanations. Compared to GPT-4o and Gemini-1.5, which lack AD integration, \NAME shows significant improvements. ~\cref{tab:qualitative_nux} presents qualitative examples comparing decoded outputs to ground-truth captions, demonstrating that \NAME produces coherent and accurate explanations of ego vehicle behavior and its underlying rationale.

\begin{table}[t]
\centering
\small
\resizebox{\columnwidth}{!}{
\begin{tabular}{@{}p{0.97\columnwidth}@{}}
\toprule
\textbf{Driving Explanation and Reasoning
% : Sample Predictions on Nu-X
} \\
\midrule
\texttt{\textcolor{teal}{\textbf{Decoded Narration:}} the car continues straight along the road }\\
\texttt{\textcolor{teal}{\textbf{Decoded Reasoning:}} because the path ahead is clear of any obstacles or traffic} \\
\texttt{\textcolor{orange}{\textbf{GT Narration:}} the car confidently advances on the open road} \\
\texttt{\textcolor{orange}{\textbf{GT Reasoning:}} as the space ahead is clear, allowing for uninterrupted progression} \\
\midrule
\texttt{\textcolor{teal}{\textbf{Decoded Narration:}} the car moves at a constant speed on the wet street} \\
\texttt{\textcolor{teal}{\textbf{Decoded Reasoning:}} because traffic is flowing freely without any obstructions} \\
\texttt{\textcolor{orange}{\textbf{GT Narration:}} the car proceeds unwaveringly along the road} \\
\texttt{\textcolor{orange}{\textbf{GT Reasoning:}} for the traffic light ahead shows a clear signal and there are no obstacles obstructing its path} \\
\midrule
\texttt{\textcolor{teal}{\textbf{Decoded Narration:}} the car stops at the intersection} \\
\texttt{\textcolor{teal}{\textbf{Decoded Reasoning:}} because the traffic light is red, requiring the vehicle to halt} \\
\texttt{\textcolor{orange}{\textbf{GT Narration:}} the car is stationary in the middle of the bustling city street} \\
\texttt{\textcolor{orange}{\textbf{GT Reasoning:}} because the red traffic light indicates a clear need for pause, preventing any forward movement} \\
\bottomrule
\end{tabular}
}
\caption{Qualitative examples of driving behavior narration and reasoning on Nu-X. Predictions by \NAME align closely with ground-truth explanations.}
\label{tab:qualitative_nux}
\vspace{-5pt}
\end{table}

\subsubsection{3D Dense Captioning}
For object-level captioning, \NAME achieves a CIDEr score of 341.9, establishing new state-of-the-art results. Compared to Hint-AD (263.7 CIDEr) and TOD3Cap (120.3 CIDEr), \NAME delivers substantial gains across all metrics, including BLEU and ROUGE-L. These improvements stem from the perception and prediction alignment modules, which tightly couple visual tokens and agent-level motion with descriptive language. Qualitative analysis further validates these quantitative results, as shown in~\cref{tab:qualitative_3dcap}.

\begin{table}[!ht]
\centering
\resizebox{\columnwidth}{!}{
\begin{tabular}{@{}l@{}}
\toprule
\textbf{3D Dense Captioning Qualitative Results} \\
\midrule
\texttt{\textcolor{teal}{\textbf{Decoded:}} Black car about 20 meters away in front of}\\
\texttt{the ego car is not moving in the carpark area.}\\[2pt]
\texttt{\textcolor{orange}{\textbf{GT:}} Black, shiny, and sleek car about 20 meters away}\\
\texttt{in front of the ego car is not moving in the carpark area.}\\
\midrule
\texttt{\textcolor{teal}{\textbf{Decoded:}} Barrier about 13 meters away in the back} \\
\texttt{left of ego car is not moving.}\\[2pt]
\texttt{\textcolor{orange}{\textbf{GT:}} Barrier about 12 meters away in the back}\\
\texttt{left of ego car is not moving.}\\
\bottomrule
\end{tabular}
}
\caption{Qualitative examples highlighting decoded versus ground-truth (GT) descriptions in the 3D dense captioning task.}
\label{tab:qualitative_3dcap}
% \vspace{-5pt}
\end{table}

\subsubsection{Visual Question Answering}
On nuScenes-QA, we report accuracy across zero-hop (H0), one-hop (H1), and overall questions. \NAME achieves the best overall accuracy of 52.9\%, surpassing Hint-AD (50.5\%) and both GPT-4o (37.1\%) and Gemini-1.5 (35.4\%) by a large margin. The model performs particularly well on the more challenging one-hop questions, demonstrating its ability to reason over multiple cues by leveraging aligned planning and motion context. Qualitative examples illustrating the reasoning capabilities of \NAME are provided in ~\cref{tab:qualitative_vqa}.

\subsubsection{Comparison with General-Purpose Multimodal LLMs}
While GPT-4o and Gemini-1.5 offer strong general vision-language capabilities, their lack of alignment with the underlying AD stack results in inconsistent or superficial outputs. These models underperform on structured reasoning tasks such as VQA and driving explanation. In contrast, \NAME explicitly grounds language generation in intermediate AD representations, offering superior interpretability, consistency, and safety.

Across all tasks, \NAME consistently outperforms prior vision-language models and traditional AD planners, achieving state-of-the-art results in both driving safety and language reasoning. These results validate the benefit of aligning perception, prediction, and planning with language.

\begin{table}[t]
\centering
\resizebox{\columnwidth}{!}{
\begin{tabular}{@{}p{0.5\columnwidth}ccc@{}}
\toprule
\textbf{Question} & \textbf{Decoded} & \textbf{GT Answer}\\
\midrule
\texttt{There is a car that is to the front of the trailer; is it the same status as the bus?} & \texttt{yes} & \texttt{yes}\\
\midrule
\texttt{There is a moving trailer; are there any stopped buses to the back of it?} & \texttt{no} & \texttt{no}\\
\midrule
\texttt{What status is the pedestrian?} & \texttt{moving} & \texttt{moving}\\
\midrule
\texttt{How many other things are in the same status as the car that is to the back right of the stopped truck?} & \texttt{6} & \texttt{5}\\
\bottomrule
\end{tabular}
}
\caption{Qualitative examples of \NAME predictions on nuScenes-QA, demonstrating reasoning accuracy on complex visual questions.}
\label{tab:qualitative_vqa}
\vspace{-5pt}
\end{table}

%% file: sec/5_conclusion.tex
\section{Conclusion}
\label{sec:conclusion}
In this paper, we introduced \NAME{}, a unified framework that achieves holistic alignment between language models and the perception–prediction–planning (P3) stack in autonomous driving systems. By designing dedicated alignment modules—Perception Alignment (P1A), Prediction Alignment (P2A), and Planning Alignment (P3A)—\NAME{} bridges the gap between spatial reasoning and natural language generation. Extensive experiments across planning, driving explanation, 3D dense captioning, and VQA tasks demonstrate that \NAME{} significantly outperforms state-of-the-art baselines in both driving safety and multimodal reasoning.

A key advantage of \NAME{} is that all alignment mechanisms are applied only during training, introducing no additional latency or computation overhead at inference time. This design ensures real-time performance while improving grounding and consistency.

\section{Limitations} 
\label{sec:limitation}
Despite its strong performance, \NAME presents some limitations. First, it requires access to intermediate representations (e.g., track, motion, and planning tokens), which may not be available in closed-source or black-box AD systems. Second, our current implementation assumes category-specific prompt selection for training, which may limit generalization to open-ended or compositional queries. Lastly, while the alignment modules improve grounding, they introduce additional complexity to the training pipeline and may require additional tuning when transferred to new AD architectures or domains.